\documentclass[sigconf]{acmart}
\AtBeginDocument{%
  \providecommand\BibTeX{{%
    \normalfont B\kern-0.5em{\scshape i\kern-0.25em b}\kern-0.8em\TeX}}}

\usepackage{times}  %Required
\usepackage{helvet}  %Required
\usepackage{courier}  %Required
\usepackage{url}  %Required
\usepackage{graphicx,verbatim}  %Required
\usepackage{algorithm}
\usepackage[noend]{algpseudocode}

\usepackage{cuted}
\usepackage[utf8]{inputenc} % allow utf-8 input
\usepackage[T1]{fontenc}    % use 8-bit T1 fonts
\usepackage{hyperref}       % hyperlinks
\usepackage{url}            % simple URL typesetting
\usepackage{booktabs}       % professional-quality tables
\usepackage{amsfonts}       % blackboard math symbols
 
\usepackage{nicefrac}       % compact symbols for 1/2, etc.
\usepackage{microtype}      % microtypography
\usepackage{soul}
\usepackage[english]{babel}

\usepackage{url}
\usepackage{hyperref}
\usepackage{graphicx}
\usepackage{caption}
\usepackage{subcaption}
\usepackage{multirow}
\usepackage{tabularx,multicol,multirow}
\usepackage{chngcntr}
\counterwithout*{table}{section}
\counterwithout*{figure}{section}
\usepackage[hang,flushmargin]{footmisc} 

\usepackage{amssymb,amsthm,mathtools}
\usepackage[capitalize]{cleveref} % must go after other package loads

\newtheorem{theorem}{Theorem}

\newcommand\numberthis{\addtocounter{equation}{1}\tag{\theequation}}
\newcommand\bbR{\ensuremath{\mathbb{R}}} % Real numbers
\newcommand\bbE{\ensuremath{\mathbb{E}}} % Expectation
\newcommand\calS{\ensuremath{\mathcal{S}}}
\newcommand\calA{\ensuremath{\mathcal{A}}}
\newcommand\extR{\ensuremath{\overline{\mathbb{R}}}}
\newcommand\allC{\ensuremath{\mathbb{R}^{\mathcal{S}\times\mathcal{A}}}}
\newcommand\psiga{\ensuremath{\psi_\text{GA}}}
\DeclareMathOperator*{\argmax}{arg\,max}
\DeclareMathOperator*{\argmin}{arg\,min}

\newcommand{\suchthat}{\;\ifnum\currentgrouptype=16 \middle\fi|\;}

\DeclarePairedDelimiter{\norm}{\lVert}{\rVert}

\newtheorem{mydef}{Definition}

\begin{document}

\title{Smooth Imitation Learning via Smooth Costs and Smooth Policies}

\author{Sapana Chaudhary}
\affiliation{%
  \institution{Texas A\&M University}
  \city{College Station}
  \state{Texas}
  \country{USA}
}
\email{sapanac@tamu.edu}

\author{Balaraman Ravindran}
\affiliation{%
  \institution{Robert Bosch Centre for Data Science and AI}
  \institution{Indian Institute of Technology Madras}
  \city{Chennai}
  \country{India}
}
\email{ravi@cse.iitm.ac.in}

\renewcommand{\shortauthors}{Chaudhary and Ravindran}

\begin{abstract}
  Imitation learning (IL) is a popular approach in the continuous control setting as among other reasons it circumvents the problems of reward mis-specification and exploration in reinforcement learning (RL). In IL from demonstrations, an important challenge is to obtain agent policies that are smooth with respect to the inputs. Learning through imitation a policy that is smooth as a function of a large state-action ($s$-$a$) space (typical of high dimensional continuous control environments) can be challenging. We take a first step towards tackling this issue by using smoothness inducing regularizers on \textit{both} the policy and the cost models of adversarial imitation learning. Our regularizers work by ensuring that the cost function changes in a controlled manner as a function of $s$-$a$ space; and the agent policy is well behaved with respect to the state space. We call our new smooth IL algorithm \textit{Smooth Policy and Cost Imitation Learning} (SPaCIL, pronounced ``Special''). We introduce a novel metric to quantify the smoothness of the learned policies. We demonstrate SPaCIL's superior performance on continuous control tasks from MuJoCo. The algorithm not just outperforms the state-of-the-art IL algorithm on our proposed smoothness metric, but, enjoys added benefits of faster learning and substantially higher average return. 
\end{abstract}


%\ccsdesc[500]{Computing methodologies~Artificial Intelligence}
%\ccsdesc[300]{Computing methodologies~Machine Learning}



\maketitle

\section{Introduction}
\iffalse
1. Problem we wish to solve; Why is it useful/important to solve this ?
2. Why is it an interesting/challenging problem ?
3. State-of-the-art methods and their drawbacks.
4. Our solution.
\fi

A vast majority of problems of interest, including but not limited to autonomous control and robotics, are characterized by high dimensional continuous (real-valued) state and action spaces \citep{recht2019tour}. Recently, a multitude of reinforcement learning (RL) approaches (like DDPG \citep{lillicrap2015continuous}, TRPO \citep{schulman2015trust}, PPO \citep{schulman2017proximal}, SAC \citep{haarnoja2018soft}, \textit{etc}) have been proposed to solve these challenging continuous control tasks. However, these algorithms require specification of a proper cost (or reward) function for any learning to be possible \citep{abbeel2004apprenticeship}. To circumvent this requirement, and to guide exploration in high dimensional continuous state-action spaces, Imitation Learning (IL) using demonstrations has been studied extensively  \citep{schaal1999imitation,calinon2009robot,argall2009survey,ho2016generative,fu2017learning}.

\begin{figure}[h]
        \centering
        \includegraphics[width=0.55\textwidth, trim= 20 50 140 180, clip]{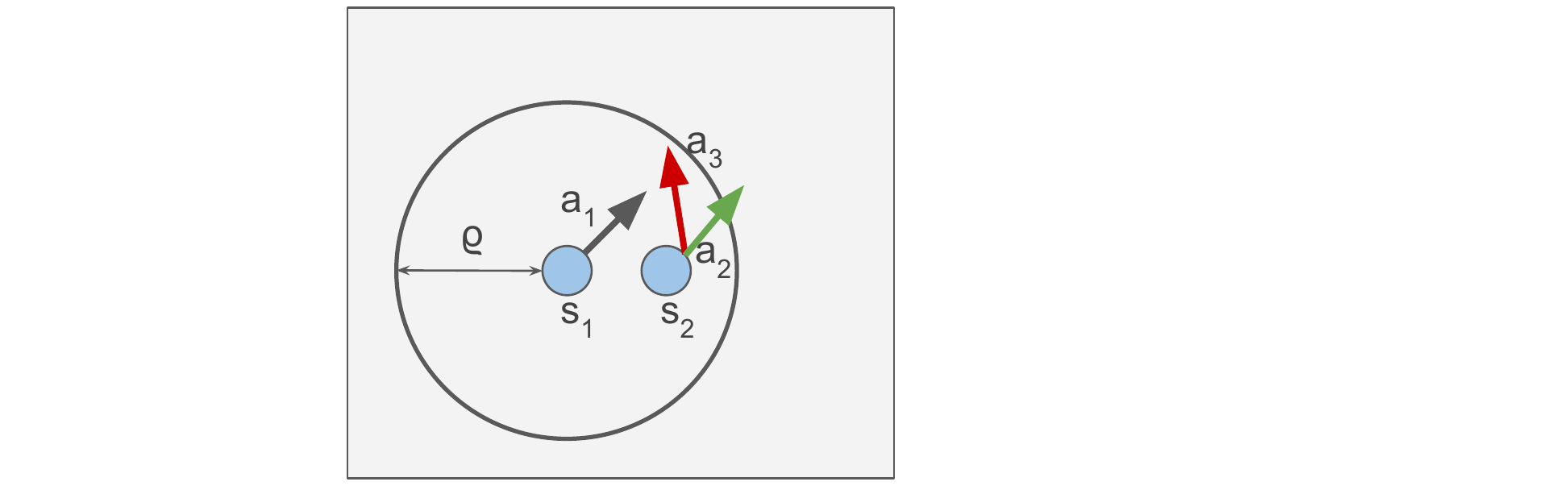}
        \caption[Understanding smoothness and non-smoothness of a policy]
    {Understanding smoothness and non-smoothness of a policy. Consider the grey region to be a portion of the environment. We use blue circles to denote the environment states, and arrows denote the actions. Let state $s_1$ be perturbed to another state $s_2$ within a radius $\varrho$ of $s_1$. Action $a_2$ comes from a smooth policy as its direction and magnitude are similar to $a_1$. On the other hand, $a_3$ comes from a non-smooth policy because of significant deviation from $a_1$ in both magnitude and direction.} \label{fig:smooth_policy}
\end{figure}

In high dimensional continuous control, an important challenge is to obtain an input-robust agent policy, \textit{i.e.}, a policy that varies in a controlled manner with respect to changes in the input state-action pair (see Fig. \ref{fig:smooth_policy}). This view of policy smoothness is equivalent to  Lipschitz-continuity of the learned agent policy. Such a smooth policy can ensure agent safety, reduce agent energy consumption, and make agent behavior predictable in desired situations \citep{kormushev2013reinforcement}.  From a control-theoretic perspective, policy smoothness certifies input-output stability (\textit{i.e}, finite $\mathcal{L}_2$ gain) during both exploration and deployment \citep{jin2018control}. In certain environments, smoothness is desired from a visual standpoint, \textit{e.g.}, smooth autonomous camera control while recording a live basketball match \citep{chen2016learning}.

While policy smoothness has recently gained attention in the RL community \citep{shen2020deep}, it has been largely untouched in IL settings. To the best of our knowledge, policy smoothness with respect to the inputs has not been previously characterized and studied in IL. It is important to note here that by virtue of learning from demonstrations (and by employing a cost (or reward) recovery scheme), IL provides us with an additional degree of control over the final agent policy. This additional control is not available in RL. To address the challenge of obtaining smooth agent policies in high dimensional continuous environments via IL using demonstrations, we propose a smooth IL algorithm: \textit{Smooth Policy and Cost Imitation Learning (SPaCIL)}. SPaCIL learns smooth agent policies by regularizing both the cost and the policy optimization steps of adversarial imitation learning setup. Adversarial IL is a state-of-the-art IL framework that uses Inverse Reinforcement Learning (IRL) approach \citep{russell1998learning,ng2000algorithms}, and recovers both a cost function and an agent policy. We demonstrate that our dual regularization of parameterized function approximators (for the cost and the policy) can assure the desired smoothness. At a higher level, the regularizers work by penalizing drastic changes in the cost and the policy at each step of the agent learning process. The policy regularization step is essential to control the smoothness of the learned policy model directly. The need for cost regularization stems from the observation that a cost function is a succinct definition of a task, and by imposing proper structure on costs, we can not only recover better costs but guide our policy optimization step towards desirable policies. It is important to note that our regularizers are not tied to a specific IL objective but are general entities that can be used with wide array of algorithms. Additionally, as a byproduct of our goal, we show that we achieve considerable gains in the training and performance of the agent policy.

\paragraph{Related Work.} A few recent works \cite{kim2004autonomous, ko2007gaussian, shen2020deep, miyato2018virtual,zhang2019theoretically,hendrycks2019using,xie2019unsupervised,jiang2019smart, blonde2020lipschitzness,le2016smooth} address some or the other notion of smoothness in adjacent/orthogonal problem settings. However, none of these works deal with smoothness in IL. The work by Kim et al. \citep{kim2004autonomous} on using reinforcement learning (RL) for the autonomous helicopter flight uses quadratic penalties for actions to modify the overall cost to encourage small actions and smooth control of the helicopter. That work deals with a low dimensional finite action setting and explicitly weights the cost associated with each action by a fixed factor. Such a weighing is not possible in high dimensional $s$-$a$ spaces. For learning controllers for autonomous blimps, Ko et al. \citep{ko2007gaussian} parameterize their controller using the slope of a policy-smoothing function that determines the controller's smoothness near a certain switching curve. However, a definite expression for policy-smoothing function exists, and the smoothness is desired over a limited region of the state space. This setting is in stark contrast to our setting, where we desire smoothness over the entire state space. Le et al. \cite{le2016smooth} study the problem of smoothly imitating an expert behaviour (using structured prediction) by ensuring that actions for adjacent states along a trajectory are similar, irrespective of proximity of adjacent states in the state space. In contrast, our notion of smoothness is not conditioned on the trajectory information and we work with the idea of smoothness of the policy space with respect to the state space. The works by Blonde et al. \citep{blonde2020lipschitzness} and Shen et al. \citep{shen2020deep} are closest to our goal. Blonde et al. \cite{blonde2020lipschitzness} show that enforcing Lipschitz-continuity of the learned reward function is essential for off-policy imitation learning to work well.  Shen et al. \cite{shen2020deep} discuss a policy regularizer to obtain smooth policies in Reinforcement Learning (RL). We, on the other hand, focus on obtaining smooth policies in IL. Additionally, Shen et al.'s \cite{shen2020deep} work does not provide a proper evaluation of smoothness. Smoothness-inducing regularizers, similar to Shen et al.'s \cite{shen2020deep}, have been previously discussed in the context of semi-supervised learning, unsupervised domain adaptation, and harnessing adversarial examples \citep{miyato2018virtual,zhang2019theoretically,hendrycks2019using,xie2019unsupervised,jiang2019smart}. State-of-the-art IL algorithm, GAIL \citep{ho2016generative} introduces various cost function regularizers to obtain various instances of IL algorithms; however, none of those regularizes enforce any special structure on the costs.

In what follows, we describe our smooth IL approach in greater detail. The main contributions of the paper are as follows:
\begin{enumerate}
    \item Formalization of the notion of smooth policies using Lipschitz continuity.  
    \item Theoretical study of how Lipschitz continuous rewards facilitate in obtaining Lipschitz continuous agent policy in on-policy continuous control. 
    \item Introduction of smoothness inducing cost function and policy function regularizers to realize a smooth IL algorithm.
    \item Introduction of a novel metric (that captures Lipschitz continuity of a learned policy) to assess the smoothness of a learned policy. % Should I talk about the quality of imitation at all?
    \item Empirical testing of smoothness of the learned policies, and validation of other claims on high dimensional continuous control tasks. 
\end{enumerate}

\section{Background}
\label{sec:background}
\paragraph{Markov Decision Process.} We consider gamma discounted infinite horizon continuous Markov Decision Processes (MDPs) \citep{sutton1998introduction} as the core framework. An MDP is specified by the tuple $<\mathcal{S}, \mathcal{A},P, c,$ $\gamma, \rho_0>$, where $\mathcal{S} \subset \mathbb{R}^{D_s} $ and $\mathcal{A} \subset \mathbb{R}^{D_a}$ are compact sets with non-zero Lebesgue measure. $\mathcal{S}$ is the set of possible states, $\mathcal{A}$ is the set of possible actions and $c: \mathcal{S} \times \mathcal{A} \rightarrow \mathbb{R}$ is the cost function. $D_s$ and $D_a$ are the dimensions of the state and the action spaces, respectively. $P$ for any $(s,a,s') \in \mathcal{S}\times\mathcal{A}\times\mathcal{S}$ triplet gives the probability of moving from state $s$ to state $s'$ on taking action $a$ at $s$. $\gamma$ is the discount factor and $\rho_0$ is the initial state distribution. 

\paragraph{Policy, occupancy measure, and expected cost.} A stationary stochastic policy, $\pi(a|s)$ gives agent's behaviour in the environment. Return, $G$, is a measure of goodness of a policy and is defined as $G(\pi) = -\mathbb{E}_{\pi}[c(s,a)] = -\mathbb{E}[\sum_{t \geq 0}\gamma^{t} c(s_t, a_t))]$. The return is estimated from trajectories as: $\hat{G}(\pi) = -\mathbb{E}_{\tau}[\sum_{t=0}^{T}\gamma^{t}c(s_t, a_t)]$, where $\tau \sim \pi$ is a trajectory of the form $\{(s_i, a_i)\}_{i=1}^T$. Here $s_0 \sim \rho_0$ is the starting state, $a_t \sim \pi(\cdot|s_t)$ and $s_{t+1} \sim P(\cdot|s_t, a_t)$. $T$ denotes the time step until which we sample a trajectory. There is a one-to-one mapping between a policy, $\pi$ and its occupancy measure, $\rho_\pi : \mathcal{S}\times\mathcal{A}\rightarrow\bbR$ defined as $ \rho_\pi(s,a) = \pi(a|s) \sum_{t=0}^\infty \gamma^t P(s_t=s|\pi)$. Expected cost in terms of $\rho_{\pi}$ is given by $\mathbb{E}_\pi[c(s,a)] = \int_{\calS\times \calA}\rho_\pi(s,a)c(s,a)~dsda$. The stationary distribution of the policy $\pi$ is denoted by $\rho_{\pi}(s)$. The expert policy is denoted by $\pi_E$. For a general function of states, $f(s)$, $\mathbb{E}_{s \sim \rho}[f(s)]$ means a $\gamma$-discounted expectation (as is for $G(\pi)$), unless stated otherwise.

\paragraph{Value functions and advantage.} The value function $V^{\pi}$ and action value function $Q^{\pi}$ can be written as $V^{\pi}(s)=-\mathbb{E}_{\pi}\left[c(\cdot, \cdot) \mid s_{0}=s\right]$ and $Q^{\pi}(s, a)=-\mathbb{E}_{\pi}\left[c(\cdot, \cdot) \mid s_{0}=s, a_{0}=a\right]$. Advantage function $A^{\pi}(s, a)=Q^{\pi}(s, a)-V^{\pi}(s)$ reflects the expected additional cost that the agent bears after taking action $a$ in state $s$.  

\paragraph{Parameterized representations.} We use parameterized function approximators (neural networks) for realization of the cost and the policy. When needed, we denote the cost function as $c_{\omega}(s,a)$ and the policy as $\pi_{\theta}(a|s)$, where $\omega \in \Omega\subset \mathbb{R}^{D_{\omega}}$ and $\theta \in \Theta\subset \mathbb{R}^{D_{\theta}}$. $D_{\omega}$ and $D_{\theta}$ are the dimensions of the parameters. We assume that $\Omega$ and $\Theta$ are compact sets. $\mathcal{C}$ is the space of all cost functions. $\Pi$ is the space of all policies. $\norm{\cdot}$ denotes the $L^2$ norm unless specified otherwise. $|\cdot|$ represents the usual norm.

\begin{comment}
GAIL defines an IRL primitive procedure, which finds a cost function such that the expert
performs better than all other policies. The cost is regularized by  $\psi : \allC \rightarrow \extR$, a closed proper convex cost function regularizer. The policy is regularized by the gamma discounted causal entropy $H(\pi) \triangleq \mathbb{E}_{\pi}[-\log \pi(a|s)]$.
\end{comment}

\paragraph{Imitation Learning.} IL solution approaches can be broadly divided into: Behaviour Cloning (BC) \citep{bain1995framework, saksena2019towards} and imitation learning using inverse reinforcement learning (IRL, \cite{abbeel2004apprenticeship,ziebart2008maximum,finn2016guided,levine2011nonlinear}). For high dimensional continuous control tasks, IL using IRL is the method of choice as BC suffers from covariate shift errors \citep{ross2010efficient, ross2011reduction}. IL using IRL can be cast as the following bi-level optimization problem (GAIL, \cite{ho2016generative}):
\begin{align}
\label{eq:irl}
\text{IL} (\pi_E) = \argmax_{c\in\mathcal{C}} &-\psi(c) + \nonumber\\ & \left(\min_{\pi\in\Pi} \bbE_{\pi} [c(s,a)]\right) - \bbE_{\pi_E} [c(s,a)],
\end{align}
where $\psi : \mathcal{C} \rightarrow \extR$ is a closed proper convex cost function loss specifier, \textit{i.e.}, it helps specify a trainable loss for the cost function model. In the first level of optimization, we fix the cost function and update the policy using $\min_{\pi\in\Pi} \bbE_{\pi} [c(s,a)]$. Next, we fix the policy from the previous update, and update the cost function using $\argmax_{c\in\mathcal{C}} -\psi(c) + \bbE_{\pi} [c(s,a)] - \bbE_{\pi_E} [c(s,a)]$.

\paragraph{Policy optimization.} The policy update step (performed using the trust region based policy optimization algorithm (TRPO, \cite{schulman2015trust})) is given by:
\begin{align}
\pi_{k+1} = \underset{\pi}{\arg\min}-\mathbb{E}_{s \sim \rho^{\pi_{{k}}}, a \sim \pi_{{k}}}\left[\frac{\pi(a \mid s)}{\pi_{{k}}(a \mid s)} A^{\pi_{{k}}}(s, a)\right] \nonumber\\
\operatorname{\textit{subject to}} ~\mathbb{E}_{s \sim \rho^{\pi_{{k}}}}\left[\mathcal{D}_{\mathrm{KL}}\left(\pi_{{k}}(\cdot \mid s) \| \pi_{}(\cdot \mid s)\right] \leq \delta.\right. \label{eq:TRPO}
\end{align}
\noindent Here, $\mathcal{D}_{\mathrm{KL}}\left(\pi_{{k}}(\cdot \mid s) \| \pi(\cdot \mid s)\right)= \mathbb{E}_{a \sim \pi_{{k}}} \left[\frac{\pi_{{k}}(\cdot \mid s)}{\pi(\cdot \mid s)}\right]$ is the Kullback–\\Leibler (KL) divergence and the constraint (in Eqn. \ref{eq:TRPO}) bounds the KL-divergence between two consecutive policies by $\delta$. Here, the expectation in constraint is not $\gamma$-discounted.

\paragraph{Cost optimization.} Using the cost function loss specifier $\psiga(c)$ of GAIL \citep{ho2016generative} the cost parameters are updated as: 
\begin{align}
c_{k+1} = &\underset{c}{\operatorname{\arg\max}}~~\mathbb{E}_{s \sim \rho^{\pi_{k}}}[\log (D(s, \pi_{k}(\cdot\mid s))]+ \nonumber \\
& \mathbb{E}_{s \sim \rho^{\pi_E}}[\log (1 - D(s, \pi_{E}(\cdot\mid s))] \label{eq:cost_opti}
\end{align}
where $\log (D(s, \pi_{k}(\cdot\mid s)) = c(s, \pi_{k}(\cdot\mid s))$ is the cost function and $D : \mathcal{S} \times \mathcal{A} \rightarrow (0,1)$ is the classifier.

\paragraph{Gaussian policy representation.} Additionally, to include stochasticity \citep{papinisafe} in the on-policy \citep{sutton1998introduction} policy gradient approach (TRPO) we consider our stationary stochastic policies to be Guassian distributed with $\mu_{\theta}(s)$ as the mean and standard deviation (std) given by a fixed quantity $\sigma$. Here, $\mu_{\theta}(s)$ is a deterministic function of the states parameterized by $\theta$. Thus, an action $a$ sampled from our policy can be written as,
\begin{multline}
    a \sim \mathcal{N}(\mu_{\theta}(s),\sigma) \implies a= \mu_{\theta}(s) + \sigma z, \quad z \sim \mathcal{N}(0,1),
    \label{eq:stoc_pol}
\end{multline}
where $\mathcal{N}(\cdot)$ is the standard Normal distribution.

\section{Problem Definition}
\subsection{Defining smooth policy, smooth cost, and smooth transition model}
This section formally defines a smooth policy, a smooth cost, and a smooth transition model.

\begin{mydef}[Smooth Policy]
Let $\xi$ be a metric on the space of policies, $\Pi$. A stationary stochastic policy $\pi(a|s): \mathcal{S}\times\mathcal{A}\rightarrow [0,1]$ is smooth with respect to the inputs, $\mathcal{S}$, if for all $s_1, s_2 \in \mathcal{S}$ it is Lipschitz continuous and hence, there exist an $M_{\pi} \geq 0$ such that
\begin{align}
    \forall s_1, s_2 \in \mathcal{S}, ~~\xi(\pi(\cdot \mid s_1), \pi(\cdot \mid s_2)) \leq M_{\pi} \norm{s_1 - s_2}. \label{eq:smooth_policy_def}
\end{align}
\end{mydef}

\begin{mydef}[Smooth Cost]
A cost function $c: \mathcal{S} \times \mathcal{A} \rightarrow \mathbb{R}$ is smooth with respect to the inputs, $\mathcal{S} \times \mathcal{A}$, if for all $(s_1, a_1), (s_2,a_2) \in \mathcal{S} \times \mathcal{A}$ it is Lipschitz continuous and hence, there exist an $M_c \geq 0$ such that
\begin{align}
    \forall (s_1, a_1), (s_2, a_2) \in \mathcal{S}, \nonumber \\ \norm{(c(s_1, a_1)-c(s_2, a_2))} \leq M_c (\norm{s_2 - s_1} + \norm{a_2 - a_1}). \label{eq:smooth_cost_def}
\end{align}
\end{mydef}

\begin{mydef}[Smooth Transition Model] \label{eq:smooth_transition_model}
If the transition model, $P$ is $L_p$-Lipschitz continuous it satisfies the following constraint for every two state action pairs $(s_1,a_1)$ , $(s_2,a_2)\in \calS\times\calA$ and all $1$-Lipschitz value functions $V$:
\begin{align}
\left|\int_{s' \in \calS}{(P(s'|s_1,a_1)-P(s'|s_2,a_2))V(s')ds'}\right|\\ \leq L_p(\|s_1 - s_2\| + \|a_1 - a_2\| ). \label{eq:smooth_transition_p}
\end{align}
\end{mydef}

If $V$ is $L_v$ Lipschitz, the right hand side of the inequality in Eqn. \ref{eq:smooth_transition_p} will be scaled by $L_v$.

\subsection{IL with smooth policy}\label{subsec:problem_definition}
Smooth policies are crucial in high dimensional continuous control for diverse reasons ranging from critical (robot safety) to aesthetic (visual appeal). When a cost function can be appropriately constructed and specified by a problem designer, reinforcement learning is the go-to solution approach. However, cost function design is a tedious task, and cost functions tend to be grossly mis-specified \citep{abbeel2004apprenticeship, dayan2002reward}. Imitation learning helps overcome the challenge of cost function design by defining strategies that learn agent policies from (expert) demonstrations. However, existing IL approaches do not guarantee that the learned policies are smooth, \textit{i.e.}, there is no existing approach that solves the following (general) problem: 
\begin{align}
    \pi^{\star}_{\text{smooth}} &= \arg \min_{\pi \in \Pi}~~ \mathbb{E}_{\pi} \left[c(s,a)| \{(s_E,a_E)\} \sim \pi_E \right] \nonumber\\
    & \text{subject to}~~ \xi \left(\pi, \pi_{\text{perturbed}} \right) \leq \epsilon^\prime, \label{eq:main_problem} 
\end{align}
where $\mathbb{E}_{\pi} \left[c(s,a)| \{(s_E,a_E)\} \sim \pi_E \right]$ represents a general IL objective, $\pi_{\text{perturbed}}$ represents a policy obtained by perturbing the original policy ($\pi$) by a small amount, and $\epsilon^\prime$ captures desired policy smoothness. The policy perturbation can be achieved in numerous ways. For our purpose of policy smoothness, $\pi_{\text{perturbed}}$ is obtained by inducing controlled amount of noise in the states along trajectories sampled by a policy $\pi$. Hence, the problem is to obtain optimal smooth agent policy, $\pi^\star_{\text{smooth}}$ from the class of policies, $\Pi$ using expert demonstration data ($N$ trajectories of the form: $\tau^E_i = \{(s_j,a_j)\}_{j=1}^T, i = 1,2,...N$). The policy optimization is subject to the constraint: $\xi \left(\pi, \pi_{\text{perturbed}} \right) \leq \epsilon^\prime$, $\textit{i.e.},$ choose that policy which when perturbed in a controlled manner behaves similar to the original unperturbed policy. The similarity in behaviour is characterized using the distance metric $\xi$ defined over $\Pi$.

\section{Method for Obtaining smooth policies in IL}
\label{sec:algorithms}
We take the approach of adversarial imitation learning \citep{ho2016generative,fu2017learning} to solve the problem discussed above (Eqn. \ref{eq:main_problem}). Adversarial IL is a high dimensional counterpart of IL using IRL, and is mathematically formulated as:
\begin{align}
\text{IL} (\pi_E) = \argmax_{c\in\mathcal{C}^\prime} \left(\min_{\pi\in\Pi} \bbE_{\pi} [c(s,a)]\right) - \bbE_{\pi_E} [c(s,a)],
\end{align}
where $\mathcal{C}^\prime$ is the cost function class modified by the loss function choosen to train the cost model. In this approach, the agent policy ($\pi$) and the cost function ($c$) are simultaneously learned using bi-level optimization of parameterized models. The cost function is learned using the expert demonstration data and the samples from the current agent policy model. The cost model update rule is defined in Eqn. \ref{eq:cost_opti}. The agent policy is learned using trust region-based policy optimization algorithm (Eqn. \ref{eq:TRPO}). To achieve the goal of policy smoothness, we propose to include smoothness inducing regularizers in both the policy and the cost optimization steps of adversarial imitation learning: 
\begin{align}
\text{IL} (\pi_E) = \argmax_{c\in\mathcal{C}^\prime} \left(\min_{\pi\in\Pi} \bbE_{\pi} [c(s,a)] + \mathcal{R}^\pi_{(s)}\right) \nonumber\\ - \bbE_{\pi_E} [c(s,a)] - \mathcal{R}^c_{(s,a)},
\end{align}
where $\mathcal{R}^\pi_{(s)}$ is the smoothness inducing regularizer on policy, and $\mathcal{R}^c_{(s,a)}$ is the smoothness inducing regularizer on cost function. The exact forms of these regularizers is discussed shortly. While policy regularization using a regularizer that captures the constraint in Eqn. \ref{eq:main_problem} seems like an obvious first approach, it is not immediately clear as to why we need to regularize the cost function. In the following section we describe the need for cost smoothness.

\subsection{Smooth costs assist in learning a smooth policy} \label{subsec:theorems}
A cost function is a central tool to perform learning through interaction. Using the following theorems, we show that a smooth cost function helps in obtaining smooth RL policies (, and hence in IL using IRL). We show that a Lipschitz continuous (smooth) cost function ensures resulting optimal value functions $V^\star(s)$ and $Q^\star(s,a)$ are Lipschitz continuous (smooth). We then show that if a Lipschitz continuous mapping is used to obtain $\pi^\star$ from $Q^\star$, then $\pi^\star$ is Lipschitz continuous. These results are of independent interest apart from providing a clear motivation to regularize cost function learning in our method.

\begin{theorem}[Generalization of Theorem 5.9 in \citep{pazis2011non} for the case of continuous state and action space] \label{th:smooth_v_q}
For a given MDP, if the cost function, $c(s,a)$  and the transition model, $P(s'|s,a)$ are $L_c$ and $L_p$-Lipschitz continuous, respectively, with respect to the state and the action pairs and $\gamma L_p < 1$ then\\
\noindent a) the optimal state value function $V^*(s)$ is $\frac{L_c}{1-\gamma L_p}$ Lipschitz with respect to states, and\\
\noindent b) the optimal state action value function $Q^*(s,a)$ is $\frac{L_c}{1-\gamma L_p}$ Lipschitz continuous with respect to states and actions.
\label{lem:lemma_3_3_1}
\end{theorem}
This theorem is a generalization of Theorem 5.9 in \cite{pazis2011non} for the case of continuous state and action space. The proof follows along the lines of one in \cite{pazis2011non} by replacing discrete Bellman optimality operator for $\mathcal{V}$ and $\mathcal{Q}$ with continuous counterparts. The proof is included in Appendix \ref{proof_theorem_1}.

\begin{theorem}
Let $\kappa$ be a pseudo-metric on the space of conditional state-action value functions $Q(s,\cdot)$. Let $H:Q(s,\cdot)\rightarrow \mu(s)$ be a smooth mapping that outputs (near) greedy stationary mean policies with respect to the input conditional state action value function $Q(s,\cdot)$ such that\\ 
$\|H(Q(s_1)) - H(Q(s_2))\| \leq L_{\mu} \kappa \left( Q(s_1,\cdot), Q(s_2,\cdot) \right), \forall s_1, s_2 \in \mathcal{S}$. \\ For a given $\frac{L_c}{1-\gamma L_p}$-Lipschitz continuous state action value function $Q^*(s,a)$, the policy obtained as $\mu^\star(s) = H(Q^*(s,\cdot))$ is Lipschitz continuous with respect to states. Then, the stationary stochastic policy $\pi^\star$ obtained from $\mu^\star$ as $\mathcal{N}(\mu^\star, \sigma)$ (for a fixed $\sigma$) is Lipschitz continuous with respect to states.
\end{theorem}
The proof is included in Appendix \ref{proof_theorem_2}.

\subsection{Exact forms of regularizers}
With the said motivations and goal in mind, we now discuss our regularizers' exact forms to achieve \textit{smooth} policies in IL.

\paragraph{Policy Smoothing}
The aim of the policy regularizer is to encourage the agent policy to be smooth within an $\epsilon$ neighbourhood of all the states sampled according to the current policy model. At each iteration of the policy update step, we sample N trajectories from the current policy where each trajectory is of the form $\tau_i = \{(s_i^t, a_i^t)\}_{t=1}^T$. We then nudge every state, $s$, in the sampled batch to obtain, $\Tilde{s}$, such that $\Tilde{s}$ lies in an $\epsilon$-radius ball around $s$ (\textit{i.e.}, $\Tilde{s} \in \mathbb{B}_{d}(s, \epsilon)$) and the policy, $\pi_{\theta}$ changes the most at this $s'$. The maximum policy change is defined using a suitable divergence $\mathcal{D}\left(\pi_{\theta}(s), \pi_{\theta}(\widetilde{s})\right)$. In this work, we consider $\mathcal{D}$ to be Jeffrey's Divergence: $\mathcal{D}_{\mathrm{J}}(P \| Q)=\frac{1}{2} \mathcal{D}_{\mathrm{KL}}(P \| Q)+\frac{1}{2} \mathcal{D}_{\mathrm{KL}}(Q \| P)$. This policy regularizer is similar in spirit to the one considered in \citep{shen2020deep}. At a fundamental level, the regularizer measures the local Lipschitz continuity of policy $\pi_{\theta}$ under the divergence $\mathcal{D}$ and thus limits the policy output decision change if a small perturbation is added to a certain state $s$. 
\begin{align}
   \mathcal{R}_{s}^{\pi}(\theta_k)=\mathbb{E}_{s \sim \rho^{\pi_{\theta_{k}}}} \max _{\Tilde{s} \in \mathbb{B}_{d}(s, \epsilon)} \mathcal{D}_{\mathrm{J}}\left(\pi_{\theta_k}(\cdot \mid s), \pi_{\theta_k}(\cdot \mid \tilde{s})\right). \label{eq:policy_reg}
\end{align}
Note that the term inside expectation is $\gamma$ discounted. This regularizer is now added to the policy optimization (TRPO) objective function to obtain smooth policy update rule: 
\begin{align}
\theta_{k+1} =& \underset{\theta}{\arg\min}-\mathbb{E}_{s \sim \rho^{\pi}_{\theta_{k}}, a \sim \pi_{\theta_k}}\left[\frac{\pi_{\theta}(a \mid s)}{\pi_{\theta_{k}}(a \mid s)} A^{\pi_{\theta_{k}}}(s, a)\right]+ \nonumber\\
& \lambda_{1} \mathbb{E}_{s \sim \rho^{\pi}_{\theta_{k}}} \max _{\widetilde{s} \in \mathbb{B}_{d}(s, \epsilon)} \mathcal{D}_{\mathrm{J}}\left(\pi_{\theta}(\cdot \mid s) \| \pi_{\theta}(\cdot \mid \tilde s)\right), \label{eq:final_policy_opti}\\
& \operatorname{\textit{subject to}} \mathbb{E}_{s \sim \rho^{\pi_{\theta_{k}}}}\left[\mathcal{D}_{\mathrm{KL}}\left(\pi_{\theta_{k}}(\cdot \mid s) \| \pi_{\theta}(\cdot \mid s)\right] \leq \delta.\right. \nonumber 
\end{align}

\paragraph{Cost Smoothing}
To obtain smooth costs, we propose to regularize the cost optimization step. The goal of this regularizer is to regulate the worst change in cost function within an $\epsilon$-ball around states obtained from the mixture trajectories, $\hat{\tau} \leftarrow \zeta\tau_E + (1-\zeta)\tau_i$, where $\zeta$ is a mixing parameter. The smoothness is induced for $(s, a)$ pairs sampled from the mixture trajectory to generalize smoothness across the entire state-action space. 

A few more words about trajectory mixing are warranted here. The mixing of trajectories considered in our work is reminiscent of policy mixing introduced in RL in Kakade et al. \cite{kakade2002approximately} and in IL introduced in Ross et al. \cite{ross2014reinforcement}. Policy mixing in these works aims to ensure conservative policy update at each policy iteration step. A greedy policy update based purely on \textit{approximate} state-action values has shown to affect the policy learning process adversely. A similar mixing of expert and non-expert data is used in the regularizer sampling distribution of WGAN-GP \citep{gulrajani2017improved}. The goal of data mixing in WGAN-GP is to ensure Lipschitzness property is satisfied by the gradients (of a model) for both the expert and the non-expert data. Our regularizer draws inspiration from the observations mentioned above. We mix agent and expert trajectories to ensure costs are smoothened both as a function of agent and expert $s$-$a$ pairs and not just imperfect agent data. Such mixing guarantees conservative enforcement of smoothness regularizer over cost function, rather than changing cost function drastically over iterations, making the learning algorithm unstable. Additionally, our regularizer provides a consistent learning signal to the cost model even when the agent and the expert policy supports do not overlap \citep{arjovsky2017towards,gulrajani2017improved}. The cost regularizer takes the following exact form:
\begin{align}
   \mathcal{R}_{(s,a)}^{c}(\theta)=\mathbb{E}_{s \sim \rho^{\hat{\pi}_{\theta}}} \max _{\Tilde{s} \in \mathbb{B}_{d}(s, \epsilon)} \|c(s, \pi_{\theta}(.|s)) -  c(s, \pi_{\theta}(.|\Tilde{s}))\|_2.  \label{eq:cost_reg}
\end{align}
Here, $\hat{\pi_{\theta}}$ is a mixture policy whose exact form is not required for our purposes. We only need samples, $\tau$ from this policy (which are obtained as $\hat{\tau} \leftarrow \zeta\tau_E + (1-\zeta)\tau_i$, where $\zeta$ is the mixing parameter). Using the regularizer of Eqn. \ref{eq:cost_reg}, the cost optimization problem becomes:  
\begin{align}
w_{k+1} = &\underset{w}{\operatorname{\arg\max}}~\mathbb{E}_{s \sim \rho^{\pi_{\theta_k}}}[\log (D_w(s, \pi_{\theta_k}(\cdot\mid s))]+ \nonumber \\
& \mathbb{E}_{s \sim \rho^{\pi_{E}}}[\log (1 - D_w(s, \pi_{\theta_E}(\cdot\mid s))] - \label{eq:final_cost_opti}\\
&\lambda_{2} \mathbb{E}_{s \sim \rho^{\hat{\pi}_{\theta}}} \max _{\Tilde{s} \in \mathbb{B}_{d}(s, \epsilon)} \|c(s, \pi_{\theta}(.|s)) -  c(s, \pi_{\theta}(.|\Tilde{s}))\|_2. \nonumber 
\end{align}

\section{Practical algorithm} \label{sec:practical_algorithm}
\paragraph{Maximization over $\epsilon$-Ball}
Both the regularized optimization problems in Eqn. \ref{eq:final_policy_opti} and Eqn. \ref{eq:final_cost_opti} require us to solve a maximization problem over the $\epsilon$-ball around a certain state $s$. The goal of this maximization problem is to find a state $s'$ within an $\epsilon$-ball of a state $s$ at which a certain function, $f(s, s')$ takes the maximum value. In Algorithm \ref{algo:reg_maximization}, we discuss a general projected gradient based approach to solve this maximization problem that is applicable to both Eqn. \ref{eq:final_policy_opti} and Eqn. \ref{eq:final_cost_opti}. For the policy smoothing regularizer of Eqn. \ref{eq:policy_reg} the $f$ in Algorithm \ref{algo:reg_maximization} is given by $f(s,s') = D_J(\pi_\theta(s),\pi_\theta(s'))$ (the Jeffrey's divergence between policies $\pi_\theta(s)$ and $\pi_\theta(s')$). For the cost smoothing regularizer of Eqn. \ref{eq:cost_reg}, $f(s,s') = \|c(s, \pi_{\theta}(.|s)) -  c(s, \pi_{\theta}(.|s'))\|_2$ (the $L_2$ distance between costs ). %The exact details of how to obtain $\nabla_{\delta} f(s, s+\delta_{\ell})$ (in step 6) and how to project onto the ball $\Pi_{\mathbb{B}_{d}}$ (step 8) is discussed in the Appendix \ref{app:projection_implementation}. 
Now that we have a procedure to obtain the regularizers, our proposed Smooth IL algorithm is provided in Algorithm \ref{algo:Safe-FW}.  

\begin{algorithm}
\caption{Maximization of $f$ over $\epsilon$-ball of a state $s$} \label{algo:reg_maximization} 
\begin{algorithmic}[1]
\State
\textbf{Input}: $s$, $\epsilon$, $\eta_{\delta}$
\State
\textbf{Initialize}: $\delta_0$\\
($N$ steps of projected gradient descent)
\State \textbf{for} $\ell = 1,2,\ldots,N$-$1$
\State \quad \textbf{(Update $\delta$ in the direction of increase in $f$)}
\State \quad $\delta_{\ell+1}=\delta_{\ell}+\eta_{\delta} \nabla_{\delta} f(s, s+\delta_{\ell})$
\State \quad \textbf{(Project $\delta_{\ell+1}$ onto the $\epsilon$-ball)}
\State \quad $\delta_{\ell+1}=\Pi_{\mathbb{B}_{d}(0, \epsilon)}\left(\delta_{\ell+1}\right)$
\State \textbf{end for} \label{line13}
\end{algorithmic}
\end{algorithm}

\begin{algorithm}
\caption{SPaCIL: Smooth Reward and Policy Imitation Learning} \label{algo:Safe-FW} %\label{alg:euclid}%($\delta , \Dc_0 , B, C, S , L , R, T,T')$
\begin{algorithmic}[1]
%\State \textbf{lization:}
%\State  \: \: $A_1 = \lambda I$
%\State  \: \: $\hat\mu_1 = 0$
\State
\textbf{Input}: Expert trajectories $\tau_{E} \sim \pi_{E}$, initial policy and discriminator parameters $\theta_{0}, w_{0}$
\State \textbf{for} $k = 1,2,\ldots,K$
\State \quad \textbf{Smooth policy update:}
\State \quad Sample $N$ trajectories $\tau_i^k \sim \pi_{\theta_k}$, $i=1,2,\cdots,N$
\State \quad Estimate regularizer in Eqn. \ref{eq:policy_reg} using Algorithm \ref{algo:reg_maximization}
\State \quad Update policy using regularized TRPO of Eqn. \ref{eq:final_policy_opti}
\State \quad \textbf{Smooth cost update:}
\State \quad Estimate regularizer in Eqn. \ref{eq:cost_reg} using Algorithm \ref{algo:reg_maximization}
\State \quad Update cost using Eqn. \ref{eq:final_cost_opti}
\State \textbf{end for} 
\end{algorithmic}
\end{algorithm}

\subsection{Evaluating smoothness of the learned policy}
To quantify the smoothness of the learned agent policy, we introduce the following (general) novel metric that captures local Lipschitz continuity of the policy:
\begin{align}
    J(\pi) = \mathbb{E}_{s\sim\rho^\pi}\left[ \max _{\Tilde{s} \in \mathbb{B}_{d}(s, \epsilon)} \mathcal{D}_{\mathrm{J}}\left(\pi(\cdot \mid s), \pi(\cdot \mid \tilde{s})\right)  \right], 
\end{align}
where the term inside expectation is \textit{not} $\gamma$-discounted. We do not include discounting here because we desire policy at any state sampled by the policy to be smooth. For the particular case of Gaussian policies, to assess the Lipschitz continuity of a stochastic policy $\pi$, we can look at the Lipschitz smoothness of its deterministic mean function $\mu$ (see Eqn. \ref{eq:stoc_pol}). For this case, practical smoothness metric takes the following form: 
\begin{align}
    J(\{\tau_i\}_{i=1}^{N}) = \frac{1}{NT} \sum_{i=1}^{N}\sum_{j=1}^{T} \max _{\Tilde{s_j} \in \mathbb{B}_{d}(s_j, \epsilon)} \frac{\|\mu(s_j) - \mu(\Tilde{s_j}) \|}{\|s - \Tilde{s_j}\|}, s_j \sim \tau_i. \label{eq:jacobian_metric}
\end{align}
We do not (on purpose) consider a global Lipschitz constant $L$ corresponding to the deterministic mean function, $\mu(s)$ of the following form:
\begin{align}
    L = \max_{s_1 \neq s_2} \frac{\|\mu(s_1) - \mu(s_2) \|}{\|s_1 - s_2\|}, \quad s_1, s_2 \in \mathcal{S}. \label{eq:metric_impractical}
\end{align}
The choice to quantify smoothness using local Lipschitz constant stems from the fact that finding the maximum over $\mathcal{S}$ in Eqn. \ref{eq:metric_impractical} is impractical for high dimensional environments. %As an alternative view, Eqn. \ref{eq:jacobian_metric} measures average spectral norm of the policy Jacobian at states sampled from the policy $\pi_{\theta}$. %More discussion on this and practical implementation of the metric is provided in the Appendix \ref{app:smoothness_metric}. 

\section{Experiments}
\begin{table*}[t]%[htbp]
\caption{\label{tab:eval_reward} 
Average Return (and one standard deviation) from 100 trajectories sampled from 5 different best agent models (\textit{i.e.}, five different random seeds). We observe that SPaCIL outperforms GAIL by a substantial margin, and enjoys much lesser variance in average return across different runs.}
\begin{center}
\setlength\tabcolsep{1pt}
\begin{tabular}{cccccc}
\toprule 
 &  \small{Reacher}  & \small{Hopper}& \small{Walker2d} & \small{HalfCheetah} &\small{Ant} \\
\midrule
$\mathcal{S}$ / $\mathcal{A}$ & $\mathbb{R}^{11}$ / $\mathbb{R}^2$ & $\mathbb{R}^{11}$ / $\mathbb{R}^{3}$ & $\mathbb{R}^{17}$ / $\mathbb{R}^{6}$ & $\mathbb{R}^{17}$ / $\mathbb{R}^6$ & $\mathbb{R}^{111}$ / $\mathbb{R}^{8}$ \\
%\small{Setting / Demo} & \textbf{S1} / 81.25 & \textbf{S3} / 1488.28 & \textbf{S2} / 969.71 & \textbf{S2} / 1843.75 & \textbf{S2} / 2109.80& \textbf{S2} / 1942.05 \\
\midrule
\small{Expert (non-smooth)}& ~~-4.18$\pm$1.79 & ~~3562.04$\pm$26.61 & ~~4224.34$\pm$18.79 &~~4022.09$\pm$79.72 &~~4872.84$\pm$568.69 \\
\midrule 
\small{GAIL} & -5.27$\pm$2.72 & 3326.43$\pm$872.42 & 4275.00$\pm$14.69 & 4064.12$\pm$120.30 & 4587.81$\pm$117.84 \\
%\small{SR-IL} & 45.01$\pm$28.16 & 628.47$\pm$69.36 & 3475.87$\pm$721.19 & 4177.13$\pm$14.86 & \textbf{4161.89}$\pm$138.05 & -68.59$\pm$19.17 \\
%\small{SP-IL} & -120.29$\pm$48.30 &1902.95$\pm$210.41 & 3664.39$\pm$16.63 & 4384.51$\pm$7.38 & 4152.26$\pm$115.86 &-3711.12$\pm$794.97 \\
\small{SPaCIL} & \textbf{-4.39$\pm$1.61} & \textbf{3662.71$\pm$39.27} & \textbf{4397.58$\pm$6.92} & \textbf{4141.10}$\pm$\textbf{93.84} & \textbf{4788.73$\pm$75.18} \\
\bottomrule 
\end{tabular}
\end{center}
\end{table*}

\begin{table*}[htbp]
\caption{\label{tab:smoothness_metric} 
Smoothness metric ($J$) (and one standard deviation) of the best trained models estimated over 150 randomly sampled trajectories from five different policy models. SPaCIL learns a substantially smooth policy irrespective of how smoothness encoded in expert demonstrations.}
\begin{center}
\setlength\tabcolsep{1pt}
\begin{tabular}{ccccccc}
\toprule 
&  \small{Reacher}  & \small{Hopper}& \small{Walker2d} & \small{HalfCheetah} &\small{Ant} \\
\midrule
\small{Expert (non-smooth)}& ~~1.77e-05$\pm$3.34e-06& ~~12.34$\pm$0.089 & ~~56.61$\pm$0.031 &~~23.88$\pm$11.70&~~6.4$\pm$3.81e-3 \\
%\small{TRPO-SR}& ~~-0.74$\pm$9.61& ~~0.0046$\pm$6.03e-5 & ~~0.5661$\pm$0.0031 &~~0.2388$\pm$0.1170&~~0.064$\pm$3.81e-4 \\
\midrule 
\small{GAIL} & 5.86e-5$\pm$9.79e-5 & 12.91$\pm$0.85 & 117.31$\pm$6.65 & 24.25$\pm$0.77 & 1.543$\pm$0.62\\
%\small{SR-IL} & 45.01$\pm$28.16 & 628.47$\pm$69.36 & 0.1350$\pm$0.0346 & 1.8739$\pm$1.5392 & 0.2711$\pm$0.0336 & -68.59$\pm$19.17 \\
%\small{SP-IL} & -120.29$\pm$48.30 &1902.95$\pm$210.41 & \textbf{0.0610}$\pm$0.0074 & 0.3315$\pm$0.0775 & 0.1810$\pm$0.0114 &-3711.12$\pm$794.97 \\
\small{SPaCIL} & \textbf{2.69e-5$\pm$2.02e-5} & \textbf{7.77}$\pm$\textbf{0.067} & \textbf{35.61$\pm$0.47} & \textbf{17.73$\pm$0.12} & \textbf{0.93$\pm$0.09} \\
\bottomrule 
\end{tabular}
\end{center}
\end{table*}

\begin{figure*}[h!]
\begin{center}
    \includegraphics[width=18cm,trim=35 240 20 220,clip]{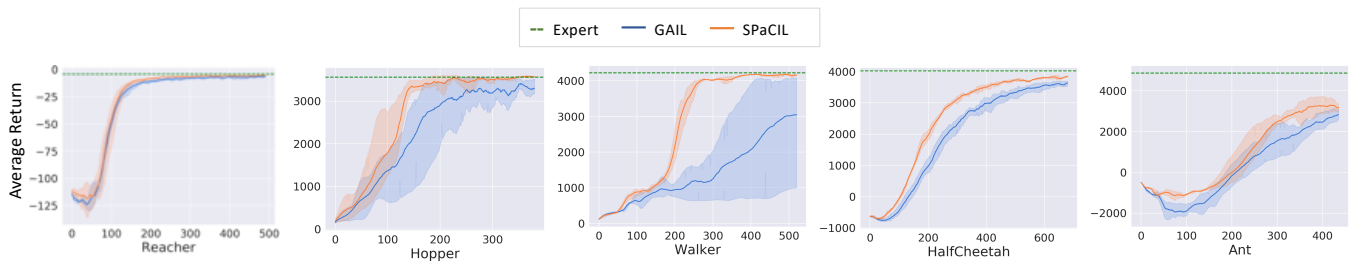}
    \caption{Learning curves for GAIL and SPaCIL on challenging continuous control tasks. The x axis is the number of algorithmic iterations. The dotted green line represents the demonstration dataset's average return. The dual regularization of SPaCIL results in faster learning.}
    \label{fig:training_curves}
    \end{center}
\end{figure*}

\paragraph{Overview}
In this section, we aim at investigating the following research questions (RQs):
\begin{enumerate}
    \item Using average return ($G$) as a metric, how well does the learned agent policy for SPaCIL and GAIL behave in the environment? \textit{(Sec. \ref{subsec:policy_avg_return_smoothness}, Table \ref{tab:eval_reward}, Fig. \ref{fig:training_curves})}
    \item Does regularization of the policy and the cost space result in faster learning for SPaCIL? \textit{(Sec. \ref{subsec:policy_avg_return_smoothness}, Fig. \ref{fig:training_curves})}
    \item Using our smoothness metric ($J$), which algorithm gives the best resulting smooth policies? \textit{(Sec. \ref{subsec:policy_avg_return_smoothness}, Table \ref{tab:smoothness_metric})}
    \item How good is our smoothness evaluation metric ($J$)? If we sufficiently perturb our smooth agent policy model, does $J$ worsen? \textit{(Sec. \ref{subsec:validating_smoothness_metric}, Fig. \ref{fig:perturbation})}
    \item How do the two regularizations affects SPaCIL's performance and in what ways? \textit{(Sec. \ref{subsec:policy_vs_cost_regularization}, Fig. \ref{fig:reg_analysis})}
    %\item Is there a trade-off between smoothness and average return? 
\end{enumerate}
We answer these questions by performing multiple experiments on continuous control tasks from MuJoCo \citep{todorov2012mujoco}. We specifically work with these environments: Reacher-v2, Hopper-v2, Walker2d-v2, HalfCheetah-v2, and Ant-v2. We would drop v2 from these environments in all future references to them.

\paragraph{Implementation details} For all the environments, the expert is trained using TRPO \cite{schulman2015trust}. We then sample trajectories from the best expert model and form our demonstration dataset. The demonstration dataset does not necessarily come from a smooth expert policy because we do not explicitly incorporate smoothness during TRPO training. This setting is practical as in reality we might have non-smooth demonstrations but would still desire a smooth controller. The imitation learning algorithms are trained on these datasets. The algorithms do not have access to the actual environment rewards and dynamics. To make the comparison fairer, for each environment, all the hyperparameters of GAIL and SPaCIL are the same except for SPaCIL's regularization-specific hyperparameters. More implementation details can be found in Appendix \ref{app:exp_setup_hyper_params}.

\subsection{Policy average return and smoothness} \label{subsec:policy_avg_return_smoothness}
As the first set of experiments, we train GAIL and SPaCIL on demonstration datasets from each environment. The algorithms are trained for 400 to 500 iterations depending on when the average return stabilizes. The learning curves are included in Fig. \ref{fig:training_curves}. The average return from SPaCIL is better than GAIL across all the environments. Higher returns mean that our smoothness regularizations result in overall better-behaved agents (RQ 1). SPaCIL also enjoys lesser variance in average return, implying the learned policies will give the said average returns with greater confidence. SPaCIL additionally performs better than the average return of the demonstration data for Hopper, Walker, and HalfCheetah.

The smoothness metric $J$ is estimated from 150 trajectories (sampled using different random seeds) from five best agent models. Here as well, SPaCIL outperforms GAIL and recovers much smoother learned agent policies (RQ 3). The training curves and $J$ for Walker are particularly interesting. GAIL's inability to tap onto the smoothness (seen from very high $J$ for this task) results in a much higher variance of the training curve. From the nature of the learning curve for Reacher, we see that the advantage of SPaCIL over GAIL is more pronounced in high-dimensional tasks. From Fig. \ref{fig:training_curves} it is evident that SPaCIL results in faster stabilized returns (RQ 2).

\subsection{Validating smoothness metric} \label{subsec:validating_smoothness_metric}
\begin{figure}[h!]
    \centering
    \includegraphics[width=8cm,trim=15 150 20 140,clip]{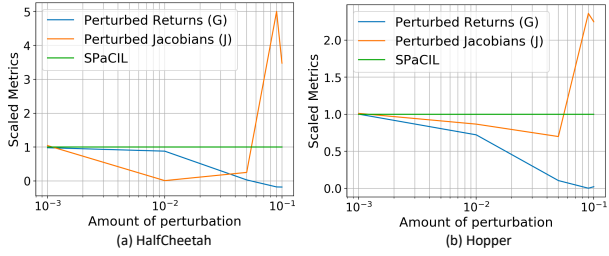}
    \caption{Variation of scaled average policy smoothness metric (J) and scaled average return for perturbed SPaCIL models for HalfCheetah and Hopper. Scaling is performed w.r.t. SPaCIL J and G, respectively. The x-axis quantifies perturbation as the standard deviation of noise added to model parameters. For sufficiently perturbed models, $J$ is very high, validating $J$ as a smoothness metric.}
    \label{fig:perturbation}
\end{figure}
This section aims to answer if the proposed smoothness metric ($J$) is truly meaningful. To answer this, we take the best SPaCIL policy model for HalfCheetah and Hopper, and perturb this model by adding a zero mean, fixed standard deviation Gaussian noise to the model parameters. Standard deviations considered are: $0.001, 0.01, 0.005, 0.009$ and $0.1$. We then obtain the average return, and the average smoothness metric ($J$) from all these perturbed models. Fig. \ref{fig:perturbation} depicts scaled average returns and average $J$ for HalfCheetah and Hopper. The scaling is done with respect to SPaCIL model's average return and average $J$ (in green colour), \textit{i.e.}, in Fig. \ref{fig:perturbation} SPaCIL model's both average return and average $J$ are equal to one. We observe that the average return decreases by a small amount with small perturbation with severe decrease when perturbation standard deviation is 0.1 (high). This means that our learned policy model is quite robust to model parameter perturbation. The smoothness metric initially decreases showing existence of a policy that is smoother than the baseline at the cost of decrease in return. This highlights an important facet of our method - we want high performing policies to be smooth, and not desire excessive smoothness at the cost of lesser return. Also, an extremely smooth policy might mean the agent can hardly move and hence gets a low return. When the model is sufficiently perturbed (with a perturbation standard deviation of 0.1), the resulting $J$ is very high, showing that the policy model is non-smooth (RQ 4).

\subsection{Policy regularization vs. cost regularization: Which is more important?} \label{subsec:policy_vs_cost_regularization}
\begin{figure}[h!]
    \centering
    \includegraphics[width=9cm,trim=150 120 180 110,clip]{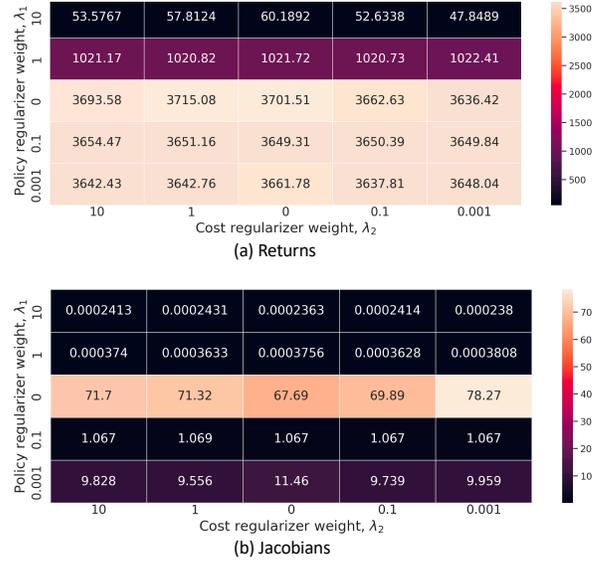}
    \caption{Average return ($G$) and average smoothness metric ($J$) variation with varying policy and cost regularization parameters ($\lambda_1$ and $\lambda_2$) for Hopper. Both policy and cost regularizations are needed to obtain high performing smooth policies.}
    \label{fig:reg_analysis}
\end{figure}

We run Hopper with varying amounts of cost and policy regularizations. Fig. \ref{fig:reg_analysis} depicts the results from this experiment. We get the best smooth policy for $\lambda_1=0.001$ and $\lambda_2=1$ with $G=3642.76$ and $J=9.556$. %Since the rows of Fig. \ref{fig:reg_analysis}(b) consist of similar values, it implies that $\lambda_1$ (the policy regularizer) swings the smoothness metric, $J$. 
The agent obtains the maximum return of $3715.08$ at $\lambda_1 = 0$ and $\lambda_2=1$, \textit{i.e.}, merely smoothing the costs results in policies that enjoy higher return. However, this policy is highly non-smooth with a $J=68.67$. Thus, the agent has to pay some additional cost to obtain visual smoothness. We can see from Fig. \ref{fig:reg_analysis} that a delicate balance of both policy and cost regularization is needed to obtain high performing smooth policies.

\section{Conclusion}
In this work, we develop a novel model-free on-policy IL algorithm: \textit{Smooth Policy and Cost Imitation Learning} (SPaCIL) that learns a smooth agent policy. Smoothness in policies is achieved by regularizing both the cost and the policy models of adversarial imitation learning framework. Through smoothness-inducing regularization, our algorithm can encode the domain knowledge about the smoothness of costs and policies. Our algorithm not only obtains smooth IL policies (measured by the policy smoothness metric that we introduce), it results in policies with a higher return than state-of-the-art adversarial IL algorithm GAIL. Our algorithm enjoys added benefits of faster learning and a much lower variance in average returns and smoothness metric.

\bibliographystyle{ACM-Reference-Format}
\bibliography{main.bib}

%\appendix

\newpage
\appendix
\section{Proofs }
\subsection{Proof of Theorem 1} \label{proof_theorem_1}
\begin{proof}
This theorem is a generalization of Theorem 5.9 in \cite{pazis2011non} for the case of continuous state and action space. The proof follows along the lines of one in \cite{pazis2011non} by replacing discrete Bellman optimality operator for $\mathcal{V}$ and $\mathcal{Q}$ with continuous counterparts. Let $\mathcal{T} \colon \mathcal{V} \rightarrow \mathcal{V}$ be the Bellman optimality operator, where $\mathcal{V}$ is the space of value functions. Then,
\begin{align}
(\mathcal{T}V)(s) = \min_a \left(c(s,a) + \gamma\int_{s' \in \calS}{p(s'|s,a)V(s')ds'}\right)
\end{align}
If $V(s)$ is $L_v$ Lipschitz in $s$, then, so is $(\mathcal{T}V)(s)$. This can be seen as follows:\\
$\forall s_1,s_2 \in \calS$ we have, 
\begin{align*}
  \left|(\mathcal{T}V)(s_1)-(\mathcal{T}V)(s_2)\right| = \nonumber\\ | \min_a \left(c(s_1,a) + \gamma\int_{s' \in \calS}{p(s'|s_1,a)V(s')ds'}\right) \nonumber\\ - \min_a \left(c(s_2,a) + \gamma\int_{s' \in \calS}{p(s'|s_2,a)V(s')ds'}\right)|. \numberthis 
\end{align*}

\noindent Define $\hat{a}$ as,
\begin{align}
\hat{a} \triangleq \begin{dcases*} \argmin_{a \in \calA}{(c(s_1,a) +
\gamma\int_{s' \in \calS}{p(s'|s_1,a)V(s')ds'})}, \\
\quad \quad \quad \text{if}~ V(s_1) \leq V(s_2) \\  \argmin_{a \in \calA}{(c(s_2,a) + \gamma\int_{s' \in \calS}{p(s'|s_2,a)V(s')ds'})}, \\
\quad \quad \quad \text{if}~V(s_1) > V(s_2).\\ \end{dcases*}
\end{align}

\noindent Then, \begin{align}
\left|(\mathcal{T}V)(s_1)-(\mathcal{T}V)(s_2)\right| \leq \nonumber \\ | c(s_1,\hat{a}) + \gamma\int_{s' \in \calS}{p(s'|s_1,\hat{a})V(s')ds'} \nonumber \\- c(s_2,\hat{a}) - \gamma\int_{s' \in \calS}{p(s'|s_2,\hat{a})V(s')ds'} |,
\end{align}
where we have used the fact that $\forall f_1, f_2 \colon \calS \rightarrow \bbR$ bounded functions such that $\min_{s \in \calS}{f_1(s)} \leq \min_{s \in \calS} f_2(s)$ and $\hat{s} = \argmin_{s \in \calS}f_1(s)$, we have $|\min_{s \in \calS} f_1(s)-\min_{s \in \calS} f_2(s)| \leq | f_1(\hat{s})- f_2(\hat{s})|$. Then using triangle inequality, we have $\forall s_1,s_2 \in \calS$,
\begin{align}
\left|(\mathcal{T}V)(s_1)-(\mathcal{T}V)(s_2)\right| \leq \|c(s_1,\hat{a}) - c(s_2,\hat{a})\| + \nonumber\\
\gamma  \left|\int_{s' \in \calS}{(p(s'|s_1,\hat{a})-p(s'|s_2,\hat{a}))V(s')ds'}\right|.
\end{align}
$(1/L_v)V$ is 1-Lipschitz, hence using the Definition \ref{eq:smooth_transition_model} and Lipschitzness of cost we get,
\begin{align}
|(\mathcal{T}V)(s_1)-(\mathcal{T}V)(s_2)| \leq (L_c + L_v L_p)\|s_1 - s_2\|.
\end{align}

\noindent Hence, $(\mathcal{T}V)(s)$ is $L_c + \gamma L_vL_p$-Lipschitz in $s$ and $a$, if $V$ is $L_v$-Lipschitz in $s$ and $a$. Choose $V_0=0$. $V^*$ is the stationary point of $\mathcal{T}$ \citep{puterman2014markov}: $V^* = \lim_{n\to\infty}\mathcal{T}^nV_0$. Hence, $V^*$ is $L_c + \gamma L_pL_c + \gamma^2L_p^2L_c + \dots = L_c/(1-\gamma L_p)$-Lipschitz continous.\\   

\noindent \textit{Proof. b)}we know that  
\begin{align}
Q^*(s,a) &=  c(s,a) + \gamma\int_{s' \in \calS}{p(s'|s,a)V^*(s')ds'}
\end{align}
For this optimal state action value function $Q^*$ and $s_1,s_2\in \calS$, we then have 
\begin{align}
|Q^*(s_1,a_1)-Q^*(s_2,a_2)| \leq |c(s_1,a_1)-c(s_2,a_2)| + \nonumber\\
\gamma \left|\int_{s' \in \calS}{(p(s'|s_1,a_1)-p(s'|s_2,a_2))V(s')ds'}\right| 
\end{align}
\begin{align}
\implies |Q^*(s_1,a_1)-Q^*(s_2,a_2)| &\leq (L_c + \gamma L_{v*}L_p)(\|s_1 - s_2\| + \|a_1 - a_2\|)
\end{align}
Using the value of $L_{v*}$ from the part a), we get 
\begin{align}
|Q^*(s_1,a_1)-Q^*(s_2,a_2)| \leq (\frac{L_c}{1-\gamma L_p})(\|s_1 - s_2\| + \|a_1 - a_2\|)
\end{align}
Hence, Proved.
%Let $\zeta$ be a metric on the space of $Q$ functions, $\mathcal{Q}$. Let $\xi$ be a metric on the space of policies, $\Pi$. Let $H:\mathcal{Q}\rightarrow \Pi$ be an operator such that for some $k>0$, $
%\left(\forall\left(Q^{\prime}, Q^{\prime \prime}\right) \in \mathcal{Q} \times \mathcal{Q}\right), \xi\left(H\left(Q^{\prime}\right), H\left(Q^{\prime \prime}\right)\right) \leq k \zeta\left(Q^{\prime}, Q^{\prime \prime}\right)$.%$H$ has an additional property of outputting (near) greedy policies with respect to the input state action value function $Q$. 
%such that $\kappa \left( Q(s_1,\cdot), Q(s_2,\cdot) \right) \leq \max_{a}|Q(s_1,a)-Q(s_2,a)|$.
\end{proof}

\subsection{Proof of Theorem 2} \label{proof_theorem_2}
\begin{proof}
From Lipschitz continuity of optimal $Q^\star$ function, we have $\forall s_1,s_2 \in \mathcal{S}$, 
\begin{align}
    \left|Q^*(s_1,a_1)-Q^*(s_2,a_2) \right| \leq \left(\frac{L_c}{1-\gamma L_p}\right)(\|s_1 - s_2\| + \|a_1 - a_2\|).
\end{align}
This can be re-written as
\begin{align}
    \left|\max_{a_1}Q^*(s_1,a_1)-\max_{a_2}Q^*(s_2,a_2)\right|\\ \leq \left(\frac{L_c}{1-\gamma L_p}\right)(\|s_1 - s_2\| + \|a_1 - a_2\|).
\end{align}

Now, it is easy to see that $|\max_{a_1}Q^*(s_1,a_1)-\max_{a_2}Q^*(s_2,a_2)| \leq \max_a |Q^*(s_1,a) - Q^*(s_1,a)|$ \cite{1742410}. From equivalence of norms, there exists a $\frac{1}{K_1}\geq0$ such that $$\max_a |Q^*(s_1,a) - Q^*(s_1,a)| \leq K_1|\max_{a_1}Q^*(s_1,a_1)-\max_{a_2}Q^*(s_2,a_2)|.$$ Note that $\max_a |Q^*(s_1,a) - Q^*(s_1,a)|$ and $\kappa$ are both pseudo-metrics on the space of $Q(s,\cdot)$ functions. Thus, from topological equivalence of pseudo-metrics, there exists a $K_2 \geq 0$ such that $\kappa \left( Q(s_1,\cdot), Q(s_2,\cdot) \right)$ $\leq K_2 \max_{a}|Q(s_1,a)-Q(s_2,a)|$. Thus, we have $\kappa \left( Q(s_1,\cdot), Q(s_2,\cdot) \right) \leq K_1 K_2 |\max_{a_1}Q^*(s_1,a_1)-\max_{a_2}Q^*(s_2,a_2)|$. Thus, for all $s_1, s_2 \in \mathcal{S}$ and for all $a \in \mathcal{A}$ we have,
\begin{align}
    \kappa \left( Q(s_1,\cdot), Q(s_2,\cdot) \right) \leq K_1 K_2 \left(\frac{L_c}{1-\gamma L_p}\right)(\|s_1 - s_2\|).
\end{align}
Now, using definition and Lipschitz continuity of $H$, we get 

\begin{align}
    \|\mu(s_1) - \mu(s_2)\| &=\|H(Q^\star(s_1,\cdot)) - H(Q^\star(s_2,\cdot))\| \\
    &\leq L_{\mu} \kappa \left(Q^\star(s_1,\cdot), Q^\star(s_2,\cdot)\right)\\
    &\leq K_1K_2L_{\mu} \frac{L_c}{1-\gamma L_p} \|s_1 - s_2\|.
\end{align}
Thus, the optimal mean policy $\mu^\star$ is $ \frac{K_1 K_2 L_{\mu}L_c}{1-\gamma L_p}$-Lipschitz continuous with respect to the states. Then, the stationary stochastic policy $\pi^\star$ obtained as $\mathcal{N}(\mu^\star(s),\sigma)$ (for a fixed $\sigma$) is $ \frac{K_1 K_2 L_{\mu}L_c}{1-\gamma L_p}$-Lipschitz continuous with respect to the states.  
\end{proof}

\section{Experimental Setup and Hyperparameter Details} \label{app:exp_setup_hyper_params}
%\subsection{The MuJoCo Tasks}

\subsection{Policy, Cost and Value Models}
%\subsubsection{Approximating factored models}
\paragraph{Policy} In our work, the policy, $\pi_{\theta}(a \mid s),$ is parameterized with a neural network. The neural network takes in a state vector $s$ and deterministically maps $s$ to a vector $\mu$. The neural network is additionally used to learn a vector $r$ of log standard deviation with the same dimension as $a .$ During training, the action is sampled stochastically from $\mathcal{N}(\mu, exp(r))$. While evaluating a policy, the mean action $\mu$ is chosen.

\paragraph{Value} The value neural network simply takes in a state $s$ and outputs scalar $V(s)$. 

\paragraph{Discriminator} The discriminator network takes in a state-action pair ($s,a$) and outputs a scalar $x$. The cost of this state-action pair is then defined as $c(s,a)= log(S(x)),$ where $S(x)=\frac{e^x}{e^x + 1}$. 

\subsubsection{Neural network details}
All the models (policy, value, and cost functions) in this work are multi-layer perceptrons (MLPs). The policy models for all the algorithms (TRPO, GAIL, and SPaCIL) have two layers of 400 and 300 neurons each. The value functions across all the algorithms have two layers of 100 neurons each. Similarly, the discriminator network (a representative for cost model) has two layers of 100 neurons each. The non-linearity used in all the models is \textit{tanh}. All the weights and biases are initialized using $U(-k,k)$ where $k=\frac{1}{\sqrt{\text{size of weights}}}$. The final layer across models has zero bias and weights multiplied by $0.1$. The step size and learning rate for the policy model is determined by the TRPO algorithm. All the other models are trained using Adam optimizer \cite{kingma2014adam} with a learning rate of $1e$-$3$ for the value network, and a learning rate of $0.01$ for the discriminator network.

\subsection{Expert learning using TRPO} \label{app:expert_learn}
The MuJoCo environments used in our experiments are from the OpenAI Gym~\cite{brockman2016openai}. We specifically work with v2 of Reacher, Hopper, Walker2d, HalfCheetah, and Ant. The expert is trained using TRPO \cite{schulman2015trust}. The TRPO hyperparameters \textit{max KL} and \textit{damping} are both fixed to $0.01$ for all the environments. The batch size is $50000$ for all the environments except Reacher's $5000$. We train all the environments for $500$ iterations except Reacher's $200$. We fix $\gamma = 0.995$ for all the environments except Reacher's $0.99$. We fix $\tau = 0.95$ for Reacher, $\tau = 0.99$ for Hopper and Walker, and $\tau = 0.97$ for the rest.

\subsection{GAIL and SPaCIL parameters} \label{app:algo_hyper_params}
The best set of hyperparameters for SPaCIL are listed in Table \ref{tbl:spacil_params}. The policy regularization strength is denoted by $\lambda_1$. The cost regularization strength is denoted by $\lambda_2$. The step size parameter in the projected gradient descent part of regularizer estimation is denoted as $\nu$ and takes a value of $0.02$ across environments. The perturbation strength around a particular state (to project onto $\epsilon$-Ball) is denoted by $\epsilon$. It takes a value of $0.01$ across environments. 
For regularization purposes, we keep the policy $\sigma$ fixed to 1. The mixing parameter $\mu$ is randomly sampled from a uniform distribution over $[0,1)$. sampled Generalized advantage estimation (GAE, \cite{schulman2015high}) parameters, $\gamma$ and $\tau$ are same as TRPO for all the environments. GAIL has all the hyperparameters, except training data size, same as that of TRPO. We are provided with a fixed number of expert trajectories before training begins. This number is the same for both GAIL and SPaCIL, and is listed in Table \ref{tbl:spacil_params}. Each trajectory consists of $50$ ($s,a$) pairs for Reacher-v2, and $1000$ ($s,a$) pairs for all the other tasks.
\begin{table}[htb!]
	\centering
	\caption{Environments details and performance of expert policies.}
	\label{tbl:spacil_params}
	\scalebox{0.86}{
		\begin{tabular}{lccccc}
			\toprule
			Environment & $\lambda_1$ & $\lambda_2$ & Expert traj No. & Agent traj No.  \\ 
			\midrule
			Reacher-v2 & $0.01$ & $0.001$ & $50$ & $50$\\
			Hopper-v2 & $0.001$ & $0.001$ & $6$ & $6$\\
			Walker2d-v2 & $0.001$ & $0.001$ & $10$ & $10$\\
			HalfCheetah-v2 & $0.01$ & $0.001$ & $6$ & $6$\\
			Ant-v2 & $0.001$ & $0.001$ & $15$ & $15$\\
			\bottomrule
		\end{tabular}
	}
\end{table}

\subsection{The MuJoCo Tasks}
\begin{figure*}[h!]
\begin{center}
    \includegraphics[width=18cm,trim=15 180 20 220,clip]{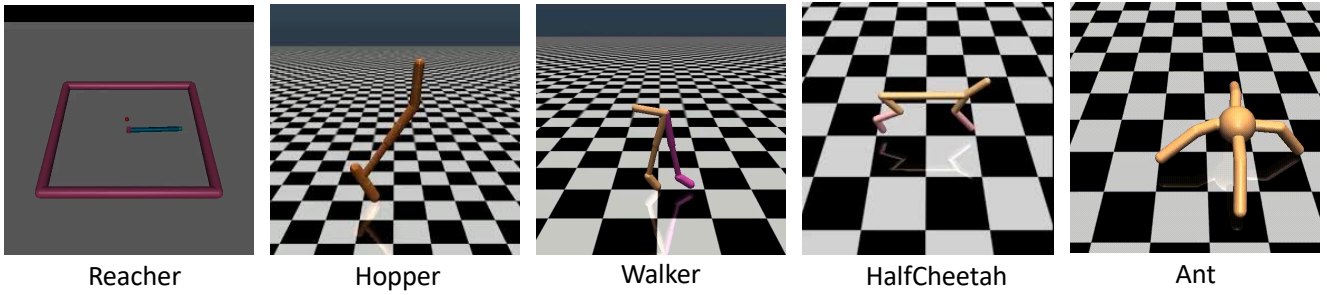}
    \caption{Robotic continuous control tasks from MuJoCo.}
    \label{fig:mujoco_envs}
    \end{center}
\end{figure*}

\paragraph{Reacher-v2.} In this environment, the agent is a grasping arm with a hinge, a body with a joint, and a tip that grasps (see Fig. \ref{fig:mujoco_envs}). With the hinge fixed at the center of a square grid environment, the agent's tip is placed at a random starting state. The agent's goal is to be able to grasp a target that is randomly spawned in the square grid.
\paragraph{Hopper-v2.} In this environment, the agent is a robot with a torso and a leg (see Fig. \ref{fig:mujoco_envs}). The agent is tasked with learning to hop through the environment. The learned walking behaviour is desired to have a stable gait.
\paragraph{HalfCheetah-v2.} In this environment, the agent is a robot with only one forelimb and one hind-limb (see Fig. \ref{fig:mujoco_envs}). The agent is tasked with learning to walk and hop through the environment. The learned walking behaviour is desired to have a stable gait.
\paragraph*{Walker2d-v2.} In this environment, the agent is a robot with a torso and two legs (see Fig. \ref{fig:mujoco_envs}). The agent is tasked with learning to jump and walk through the environment. The learned walking behaviour is desired to have a stable gait.
\paragraph*{Ant-v2.}In this environment, the agent is a robot with four legs (see Fig. \ref{fig:mujoco_envs}). The agent is tasked with learning to hop through the environment. The learned walking behaviour is desired to have a stable gait.

\subsection{Deep RL tricks used in the implementation} \label{app:deep_rl_tricks}
\begin{figure*}[h]
\begin{center}
    \includegraphics[width=18cm,trim=15 240 20 220,clip]{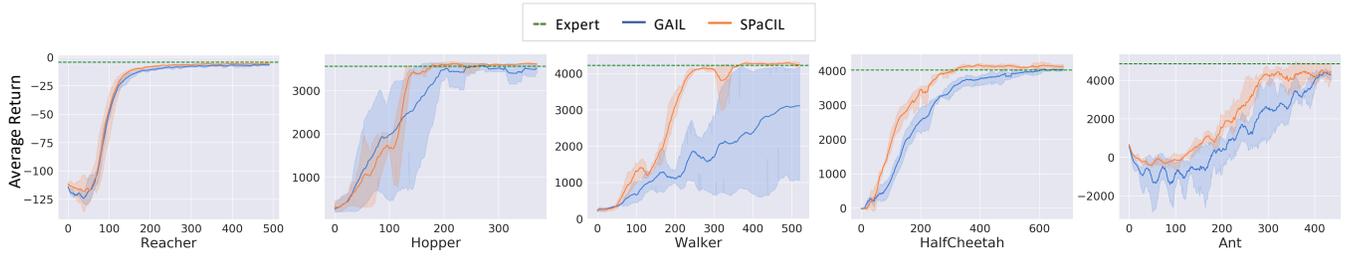}
    \caption{Evaluation curves for GAIL and SPaCIL on challenging continuous control tasks. The dotted green line represents the average return of the demonstration dataset. The dual regularization of policies and costs results in faster learning for SPaCIL.}
    \label{fig:eval_curves}
    \end{center}
\end{figure*}

We use the following standard tricks from deep reinforcement learning (DRL) in our implementation: \begin{enumerate}
    \item We keep and use a running average of the states to deal with covariate shift in the input data. 
    \item We evaluate the performance of our algorithms on a separate test environment. We keep track of the best evaluation return to save our best agent model. We sample $20,000$ $s$-$a$ pairs at each test iteration for testing. The action is sampled as the mean action ($\mu(s)$) at a certain state ($s$) rather than from the stochastic policy. The evaluation performance curves for GAIL and SPaCIL are included in Fig. \ref{fig:eval_curves}). 
    \item Different components of our code are run on different devices (\textit{i.e.}, CPU and GPU) to optimize for algorithm's run time. While sampling trajectories from the agent policy, we use multi-threading and run the code on the CPU. All the gradient calculations vis back-propagation are performed on GPU cores. 
\end{enumerate}

\begin{table*}[htbp]
\caption{\label{tab:perturbation} 
Validating our smoothness metric is a good measure of smoothness}
\begin{center}
\setlength\tabcolsep{1.2pt}
\begin{tabular}{lccccccc}
\toprule 
&  0  & 0.1 & 0.09 & 0.05 & 0.01 &0.001 \\
\midrule
\small{Reacher G}& ~~-4.18$\pm$1.79
& ~~-163.44$\pm$43.95 &~~-47.07$\pm$11.08 
&~~-30.12$\pm$10.25 &~~-11.67$\pm$2.18
&~~-12.35$\pm$2.14 \\
\small{Reacher J}& ~~2.69e-5$\pm$2.02e-5
& ~~14.42$\pm$2.06 &~~1.02$\pm$0.12 
&~~0.29$\pm$0.05 &~~8.67e-4$\pm$1.79e-4
&~~3.72e-4$\pm$7.85e-5 \\
\midrule 
\small{Hopper G}& ~~3662.71$\pm$39.27& ~~79.73$\pm$73.48 &~~4.69$\pm$0.033 & ~~383.59$\pm$313.75 &~~2644.44$\pm$1207.97
&~~3667.35$\pm$9.72 \\
\small{Hopper J}& 
~~7.77$\pm$0.067
& ~~18.24$\pm$3.01 &~~18.34$\pm$0.0162 & ~~5.43$\pm$0.13 
&~~6.732$\pm$0.27
&~~7.86$\pm$0.19 \\
\midrule
\small{Walker2d G}& ~~4397.58$\pm$6.92& ~~3.21$\pm$2.81 &~~61.92$\pm$53.04 & ~~245.21$\pm$119.8 &~~4199.56$\pm$17.19
&~~4329.44$\pm$10.69 \\
\small{Walker2d J}& ~~35.61$\pm$0.067& ~~85.91$\pm$1.06 &~~125.97$\pm$6.06 & ~~37.56$\pm$1.60 &
~~39.37$\pm$1.1
&~~38.53$\pm$0.12 \\
\midrule
\small{HalfCheetah G}& ~~4141.10$\pm$93.84& ~~-743.56$\pm$60.59 &~~-729.94$\pm$242.07 & ~~128.97$\pm$141.52 &~~3642.23$\pm$75.71
&~~4061.67$\pm$75.59 \\
\small{HalfCheetah J}& ~~17.73$\pm$0.11 & ~~61.63$\pm$6.72 &~~88.67$\pm$33.76 & ~~4.47$\pm$2.03 
&~~17.72$\pm$0.19
&~~18.51$\pm$0.21 \\
\midrule
\small{Ant G}
& ~~4788.73$\pm$75.18
& ~~-4880.08$\pm$1719.81 &~~-1454.50$\pm$961.96 & ~~1685.58$\pm$482.09 &~~4561.18$\pm$617.04
&~~4628.54$\pm$893.13 \\
\small{Ant J}
& ~~0.93$\pm$0.09& ~~186.91$\pm$74.77 &~~43.58$\pm$27.72 
& ~~1.64$\pm$2.27 
&~~0.91$\pm$0.07
&~~0.97$\pm$0.04 \\
\bottomrule 
\end{tabular}
\end{center}
\end{table*}

\section{Implementation of Projection} \label{app:projection_implementation}
At the crux of our algorithm are the following policy and cost regularizers:
\begin{align}
   \mathcal{R}_{s}^{\pi}(\theta_k)=\mathbb{E}_{s \sim \rho^{\pi_{\theta_{k}}}} \max _{\Tilde{s} \in \mathbb{B}_{d}(s, \epsilon)} \mathcal{D}_{\mathrm{J}}\left(\pi_{\theta_k}(s), \pi_{\theta_k}(\widetilde{s})\right),  \label{eq:policy_reg_appendix}
\end{align}
and 
\begin{align}
   \mathcal{R}_{(s,a)}^{c}(\theta)=\mathbb{E}_{s \sim \rho^{\hat{\pi}_{\theta}}} \max _{\Tilde{s} \in \mathbb{B}_{d}(s, \epsilon)} \|c(s, \pi_{\theta}(.|s)) -  c(s, \pi_{\theta}(.|\Tilde{s}))\|_2.  \label{eq:cost_reg_appendix}
\end{align}
Both the regularizers in Eqn. \ref{eq:policy_reg_appendix} and Eqn. \ref{eq:cost_reg_appendix} require us to solve a maximization problem over the $\epsilon$-ball around a certain state $s$. The goal of this maximization problem is to find a state $s'$ within an $\epsilon$-ball of a state $s$ at which a certain function, $f(s, s')$ takes the maximum value. In Algorithm \ref{algo:reg_maximization_appendix}, we discuss a general projected gradient based approach to solve this maximization problem that is applicable to both Eqn. \ref{eq:policy_reg_appendix} and Eqn. \ref{eq:cost_reg_appendix}. For the policy smoothing regularizer of Eqn. \ref{eq:policy_reg_appendix} the $f$ in Algorithm \ref{algo:reg_maximization_appendix} is given by $f(s,s') = D_{\mathrm{J}}(\pi_\theta(s),\pi_\theta(s'))$ (the Jeffrey's divergence between policies $\pi_\theta(s)$ and $\pi_\theta(s')$). For the cost smoothing regularizer of Eqn. \ref{eq:cost_reg}, $f(s,s') = \|c(s, \pi_{\theta}(.|s)) -  c(s, \pi_{\theta}(.|s'))\|_2$ (the $L_2$ distance between costs ). Here we provide exact details of how to obtain $\nabla_{\delta} f(s, s+\delta_{\ell})$ (in step 6) and how to project onto the ball $\Pi_{\mathbb{B}_{d}}$ (step 8).\\

\noindent For $f(s,s') = D_J(\pi_\theta(s),\pi_\theta(s'))$ and Guassian distributed policies (\textit{i.e.}, $\pi_{\theta}(\cdot \mid s) \stackrel{d}{=} \mathcal{N}(\mu_{\theta}(s),\sigma^2)$), $f$ is reduced to $\frac{\|\mu_{\theta}(s) - \mu_{\theta}(s')\|_2^2}{\sigma^2}$. Gradient of this quantity is estimated using automatic back-propagation \cite{hecht1992theory} through the policy neural network. Gradient of $f(s,s') = \|c(s, \pi_{\theta}(.|s)) -  c(s, \pi_{\theta}(.|s'))\|_2$ can be equivalently estimated using back-propagation.\\ 

\noindent Once we have $\delta_{\ell+1}$ from Step 6 of Algorithm \ref{algo:reg_maximization_appendix}, the projection onto the ball ${\mathbb{B}_{d}(0, \epsilon)}$ is evaluated using the following formula: 
\begin{align}
    \delta_{\text{new}} = \delta_{\text{old}}\text{min}\Big\{1,\frac{\epsilon}{\|\delta_{\text{old}}\|_2}\Big\}.
\end{align}

\begin{algorithm}
\caption{Maximization of $f$ over $\epsilon$-ball of a state $s$} \label{algo:reg_maximization_appendix} 
\begin{algorithmic}[1]
\State
\textbf{Input}: $s$, $\epsilon$, $\eta_{\delta}$
\State
\textbf{Initialize}: $\delta_0$\\
($N$ steps of projected gradient descent)
\State \textbf{for} $\ell = 1,2,\ldots,N$-$1$
\State \quad \textbf{(Update $\delta$ in the direction of increase in $f$)}
\State \quad $\delta_{\ell+1}=\delta_{\ell}+\eta_{\delta} \nabla_{\delta} f(s, s+\delta_{\ell})$
\State \quad \textbf{(Project $\delta_{\ell+1}$ onto the $\epsilon$-ball)}
\State \quad $\delta_{\ell+1}=\Pi_{\mathbb{B}_{d}(0, \epsilon)}\left(\delta_{\ell+1}\right)$
\State \textbf{end for} \label{line13_appendix}
\end{algorithmic}
\end{algorithm}

\section{Discussion on smoothness metric} \label{app:smoothness_metric}
\paragraph{An alternative view of smoothness metric}
To quantify the smoothness of the learned policy, we had introduced a novel metric that took the following form: 
\begin{align}
    J(\{\tau_i\}_{i=1}^{N}) = \frac{1}{NT} \sum_{i=1}^{N}\sum_{j=1}^{T} \max _{\Tilde{s_j} \in \mathbb{B}_{d}(s_j, \epsilon)} \frac{\|\mu(s_j) - \mu(\Tilde{s_j}) \|}{\|s - \Tilde{s_j}\|}, s_j \sim \tau_i. \label{eq:j_form_1}
\end{align}
where $\mu(s)$ is the deterministic mean function for $\pi(\cdot\mid s)$. We had gotten the above form for the metric by starting out by defining Lipschitz constant $L$ corresponding to the deterministic mean function, $\mu(s)$ for a general norm is given by
\begin{align}
    L = \max_{s_1 \neq s_2} \frac{\|\mu(s_1) - \mu(s_2) \|}{\|s_1 - s_2\|}, \quad s_1, s_2 \in \mathcal{S}.
\end{align}
Taking $s_1 = s + \delta s$ and $s_2 = s$
\begin{align}
    L = \max_{\delta s \neq 0} \frac{\|\mu(s + \delta s) - \mu(s) \|}{\|s + \delta s - s\|},
\end{align}
Using the first order Taylor series approximation for $\mu$ at $s$ : $\mu(s + \delta s) = \mu(s) + J_{\mu(s)} \delta s $ we get  
\begin{align}
    \label{eqn:spec_norm}
    L = \max_{\delta s_2 \neq 0} \frac{\|J_{\mu(s)} \delta s \|}{\|\delta s\|},
\end{align}
For $L^2$ norm, the quantity on the right in Eqn. \ref{eqn:spec_norm} is the spectral norm of $J_{\mu(s)}$. Hence, the local Lipschitz constant, $L$ at a particular state $s$ is given by the spectral norm of the Jacobian, $J$ at that state:  $\|J_{\mu}(s)\|_2 = \sigma_{max}(J_{\mu}(s))$ , \textit{i.e.}, the maximum singular value \cite{teel1995examples,eriksson2003applied}. Therefore, another approach to evaluate the smoothness of $\pi$, is to estimate the expected Jacobian norm using sampled trajectories as, 
\begin{align}
    \hat{E}[\|J_{\mu(s)}\|_2] = \frac{1}{NT}\sum_{i=1}^{N} \sum_{j=1}^{T} \|J_{\mu(s_j)}\|_2, \quad s_j \sim \tau_i
    \label{eq:j_form_2}
\end{align}
where $T$ and $N$ are the number of sampled trajectories and time steps, respectively. A sampled trajectory, $\tau \sim \pi$ is a trajectory of the form $\{s_0, a_0, s_1, a_1,....., s_T\}$, where $s_0 \sim \rho_0$ is the starting state, $a_t \sim \pi(\cdot|s_t)$, and $T$ denotes the time step at which we terminate an episode. The quantity in Eqn. \ref{eq:j_form_2} is the average spectral norm instead of the maximum of the norms. Hence, in our work, we use Eqn. \ref{eq:j_form_1} as the smoothness metric. $J(\{\tau_i\}_{i=1}^{N})$ in Eqn. \ref{eq:j_form_1} can be estimated using the batch $\{\tau_i\}_{i=1}^{N}$ of data sampled from any policy. 

\section{More results} \label{app:more_results}
We include results from some more experiments here. Fig. \ref{fig:eval_curves} shows the average return (and one standard deviation) over an evaluation environment. Table \ref{tab:perturbation} reports average return and average metric for perturbed models various tasks where perturbation is the variance of Gaussian noise added to the original model parameters.

\end{document}